\documentclass{article}

\usepackage{microtype}
\usepackage{graphicx}
\usepackage{subfigure}
\usepackage{booktabs} % for professional tables
\usepackage{makecell}
\usepackage{enumitem}
\usepackage{hyperref}
\usepackage{multirow}
\usepackage{xcolor}    % 
\usepackage{tcolorbox}
\usepackage{colortbl}  % 
\definecolor{lightgray}{HTML}{E3E3E3}
\usepackage[accepted]{icml2025}

\usepackage{amsmath}
\usepackage{amssymb}
\usepackage{mathtools}
\usepackage{amsthm}
\usepackage[ruled,vlined]{algorithm2e}
\usepackage[capitalize,noabbrev]{cleveref}

\theoremstyle{plain}

\theoremstyle{definition}

\theoremstyle{remark}

\usepackage[textsize=tiny]{todonotes}

\begin{document}

\twocolumn[
\icmltitle{Efficient Multi-modal Long Context Learning for Training-free Adaptation}



\icmlsetsymbol{corresponding}{$\ast$}
\begin{icmlauthorlist}
\icmlauthor{Zehong Ma}{sch}
\icmlauthor{Shiliang Zhang}{sch,lab,corresponding}
\icmlauthor{Longhui Wei}{comp}
\icmlauthor{Qi Tian}{comp,comp2}
\end{icmlauthorlist}

% \icmlaffiliation{yyy}{State Key Laboratory of Multimedia Information Processing, School of Computer Science, Peking University}
\icmlaffiliation{comp}{Huawei Inc.}
\icmlaffiliation{sch}{State Key Laboratory of Multimedia Information Processing, School of Computer Science, Peking University}
\icmlaffiliation{lab}{Peng Cheng Laboratory, Shenzhen, China}
\icmlaffiliation{comp2}{Guangdong Laboratory of Artificial Intelligence and Digital Economy
(SZ)}

\icmlcorrespondingauthor{Shiliang Zhang}{slzhang.jdl@pku.edu.cn}

\icmlkeywords{Multi-modal Long Context Learning, Training-free Adaptation, Adaptive Pruning, ICML}
\vskip 0.3in
]



\printAffiliationsAndNotice{}  % leave blank if no need to mention equal contribution

\begin{abstract}
Traditional approaches to adapting multi-modal large language models (MLLMs) to new tasks have relied heavily on fine-tuning. This paper introduces Efficient Multi-Modal Long Context Learning (EMLoC), a novel training-free alternative that embeds demonstration examples directly into the model input. EMLoC offers a more efficient, flexible, and scalable solution for task adaptation. Because extremely lengthy inputs introduce prohibitive computational and memory overhead, EMLoC contributes a chunk-wise compression mechanism combined with layer-wise adaptive pruning. It condenses long-context multimodal inputs into compact, task-specific memory representations. By adaptively pruning tokens at each layer under a Jensen-Shannon divergence constraint, our method achieves a dramatic reduction in inference complexity without sacrificing performance. This approach is the first to seamlessly integrate compression and pruning techniques for multi-modal long-context learning, offering a scalable and efficient solution for real-world applications. Extensive experiments on diverse vision-language benchmarks demonstrate that EMLoC achieves performance on par with or superior to naive long-context approaches. Our results highlight the potential of EMLoC as a groundbreaking framework for efficient and flexible adaptation of multi-modal models in resource-constrained environments. Codes are publicly available at \href{https://github.com/Zehong-Ma/EMLoC}{https://github.com/Zehong-Ma/EMLoC}.
\end{abstract}

\section{Introduction}
In recent years, multi-modal large models (MLLMs)~\cite{liu2023llava,zhu2024minigpt,qwenvl} have achieved significant advancements, demonstrating remarkable success across a wide range of multi-modal tasks. Traditionally, transferring these models to downstream tasks relies on supervised fine-tuning, including full fine-tuning and parameter-efficient methods~\cite{hu2022lora}. However, these techniques require updating model parameters and often incur substantial computational costs. Recently, MLLMs have evolved from processing single-image inputs~\cite{liu2023llava} to handling multi-image and video data~\cite{wang2024qwen2, hu2024minicpm, chen2024far}, while also supporting increasingly long context lengths.

\begin{figure}
    \centering
    \includegraphics[width=1\linewidth]{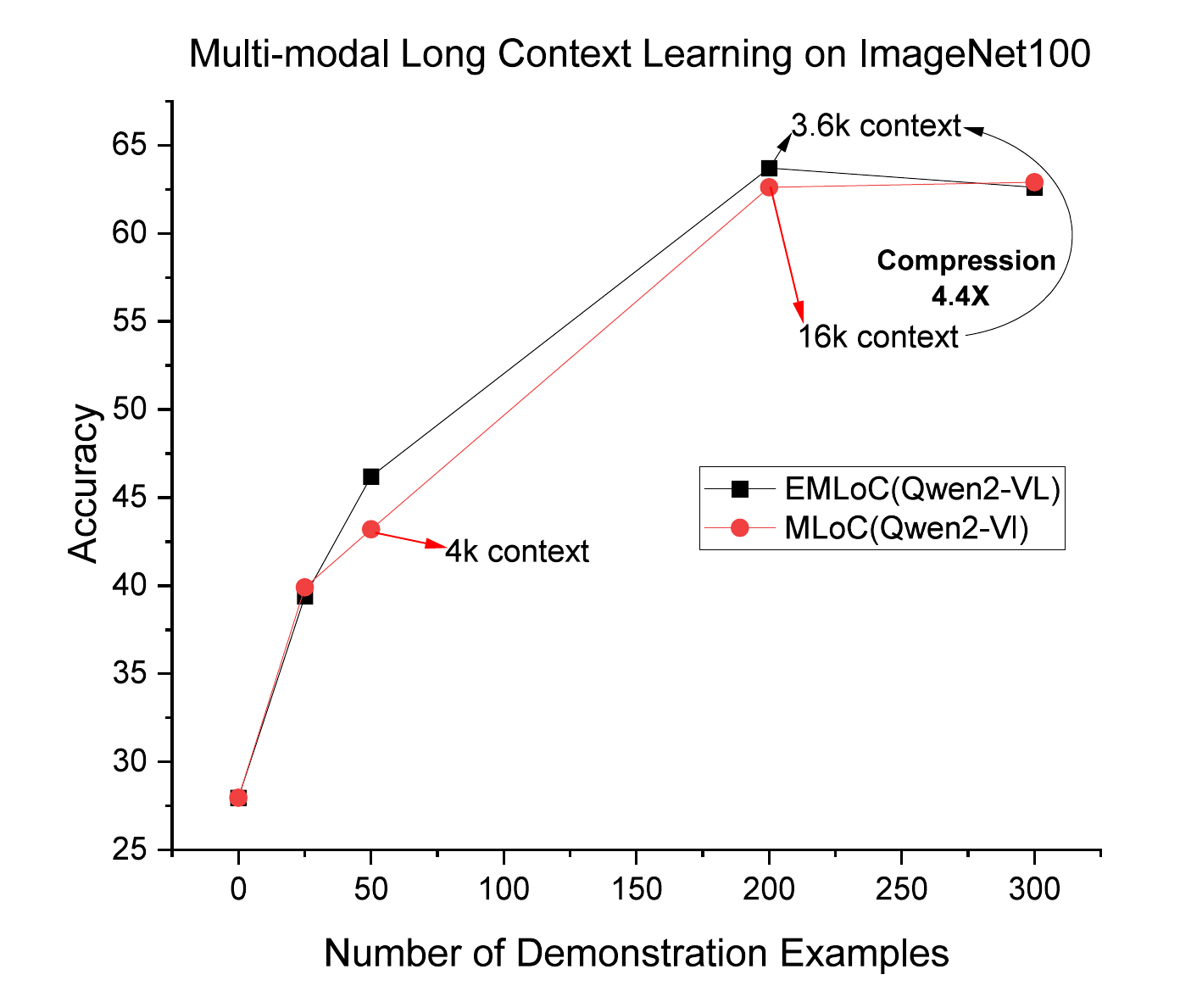}
    % \vspace{-5pt}
    \caption{The comparison between EMLoC and MLoC on ImageNet100 with varying numbers of demonstration examples. With 200 examples, EMLoC achieves 4.4× context compression over vanilla MLoC without performance loss. It significantly outperforms MLoC with 50 examples using a similar context length.}
    \label{fig:introduction}
\end{figure}

A key observation, as illustrated in~\cref{fig:introduction}, is that providing task-specific demonstration examples during inference significantly enhances model performance. We refer to this approach as Multi-modal Long Context learning (MLoC). However, directly feeding multiple multi-modal examples into the model often results in excessively long contexts, leading to prohibitively high inference costs.

To address these challenges, we introduce \textbf{E}fficient \textbf{M}ulti-modal \textbf{Lo}ng \textbf{C}ontext Learning for training-free adaptation (EMLoC), a training-free adaptation method. EMLoC leverages the benefits of multi-modal long context learning while mitigating its computational drawbacks through a chunk-wise compression mechanism combined with layer-wise adaptive pruning. This approach compresses extensive multi-modal contexts into compact, task-specific memory representations, significantly reducing computational overhead without compromising task performance.

The chunk-wise compression mechanism partitions a long context into smaller, manageable chunks, enabling a divide-and-conquer approach. This strategy reduces reliance on large-memory devices and significantly lowers computational overhead. Within each chunk, we employ a layer-wise adaptive pruning strategy, progressing from the top layer to the bottom layer. Using a greedy search algorithm, we determine the minimum number of tokens required per layer to ensure the Jensen–Shannon (JS) divergence between the original and pruned outputs remains below a predefined threshold. This approach avoids noticeable performance degradation while maximizing token reduction. Unlike static methods~\cite{xiao2024efficient,li2024snapkv,cai2024pyramidkv}, which reserve a fixed number of tokens per layer, our adaptive strategy dynamically prunes tokens based on layer-specific importance, ensuring optimal efficiency without sacrificing accuracy.

The EMLoC framework operates without modifying model weights and requires only a few forward passes to generate task-specific compressed memory. This makes it a lightweight, plug-and-play solution that maintains task performance while drastically reducing inference costs.
The contributions of this work are threefold:
\begin{itemize}
    \item Efficient Multi-modal Long Context Learning: We introduce chunk-wise compression and layer-wise adaptive pruning to reduce computational costs and memory usage in long-context scenarios.
    \item Adaptive Layer Importance Analysis: We dynamically analyze the importance of different layers in a layer-wise manner, offering new insights into token pruning strategies.
    \item We establish a linear upper bound for the information loss and demonstrate the effectiveness of our method across diverse tasks with extensive experiments.
\end{itemize}
\section{Related Work}
This work is related to multi-modal large language models, long context learning, and training-free context compression. We briefly review recent advances in these areas and discuss our contributions and differences with them.

\subsection{Multi-modal Large Language Models}
The progress in LLMs has propelled the advancement of MLLMs. Flamingo\cite{alayrac2022flamingo} pioneered the integration of a pre-trained visual encoder with the Chinchilla 70B\cite{hoffmann2022chinchilla} LLM, demonstrating strong zero-shot and few-shot performance on vision-language tasks. Since then, numerous open-source models have emerged, including Kosmos-1\cite{huang2023kosmos1}, MiniGPT-4\cite{zhu2023minigpt}, and LLaVA~\cite{liu2023llava}. Subsequent research has expanded MLLMs' functional capabilities and improved their visual perception. Models like Kosmos-2\cite{peng2023kosmos2}, CogVLM\cite{wang2023cogvlm}, Shikra\cite{chen2023shikra}, Pink\cite{Xuan_2024_CVPR}, and LocLLM\cite{wang2024locllm} incorporate localization through the pix2seq paradigm or connections with detection and segmentation models. Others, such as Qwen-VL\cite{qwenvl}, Yi-VL\cite{ai2024yi}, DeepSeek-VL\cite{lu2024deepseekvl}, InternVL\cite{chen2024far}, and Intern-XComposer\cite{dong2024internlm}, enhance capabilities with high-resolution inputs and larger training datasets. Recently, models such as Intern-XComposer-2.5~\cite{zhang2024internlm}, InternVL-2~\cite{chen2024far}, MiniCPM-V-2.6~\cite{hu2024minicpm}, and Qwen2-VL~\cite{wang2024qwen2} have advanced to support multi-image understanding, video comprehension, and multi-modal in-context learning. These recent studies have made it possible to explore MLLMs' multi-modal long-context learning.

\subsection{Long Context Learning}
In-context learning (ICL) is a good adaptation method to improve the performance of downstream tasks. We focus on multi-modal long context learning for training-free adaptation with many in-context examples.

\noindent\textbf{Scaling ICL.} The foundational study by \cite{GPT3} demonstrates that increasing the number of in-context examples improves the performance of LLMs.
% However, these experiments involved a relatively small number of examples (10 to 100), likely constrained by the limited context size, such as 2048 tokens for GPT-3. 
More recent works begin to explore the effects of scaling in-context examples beyond 1000, with \cite{li2023context}, \cite{agarwal2024many}, and \cite{bertsch2024context} showing consistent performance improvements in multiple text-based tasks. However, these studies are limited to text-only benchmarks and lack exploration in multi-modal tasks.

\noindent\textbf{Multi-modal ICL.} Recent studies have evaluated the generalization abilities of models like GPT-4V and Gemini, highlighting that in-context learning enhances performance on out-of-domain and out-of-distribution tasks \cite{zhangood, hanshift}. However, these works have not fully explored the potential of leveraging extended context windows to incorporate more demonstration examples and long-context inputs. A recent study by \cite{jiang2024manyshot} investigated many-shot in-context learning in MLLMs, revealing that open-source MLLMs struggle with complex long contexts, while closed-source models perform significantly better. This paper focuses on addressing the challenges of multi-modal long-context learning in open-source MLLMs, presenting an initial exploration of this domain. Our work aims to bridge the gap by enabling efficient and scalable long-context adaptation for open-source models.

\subsection{Training-free Context Compression}
Training-free context compression techniques have advanced significantly to address the challenges of handling extended sequences in LLMs. StreamingLLM~\cite{xiao2024efficient} retains Key-Value (KV) pairs from initial and recent tokens, leveraging attention concentration to reduce memory usage. SnapKV~\cite{li2024snapkv} dynamically identifies critical KV pairs based on attention patterns, minimizing cache size while preserving performance. H2O~\cite{zhang2023h2o} prioritizes recent and high-attention tokens with a dynamic cache strategy, optimizing memory efficiency. PyramidKV~\cite{cai2024pyramidkv} adjusts KV cache size hierarchically, allocating more memory to lower layers for efficient compression. 
Recent studies, such as LLaVolta~\cite{chen2024llavolta} and FastV~\cite{chen2025image}, have focused on compressing visual contexts to enhance the inference accuracy of multi-modal large language models. Despite these advancements in text-only and image-only context compression, multi-modal context compression remains relatively underexplored. In this work, we aim to adaptively compress multi-modal long contexts into an efficient compact memory without sacrificing performance.

\section{Method}
This section introduces our EMLoC for training-free adaptation. We begin with an overview of the approach, followed by the introduction of chunk-wise compression and layer-wise adaptive pruning. Finally, we present a theoretical analysis that establishes a linear upper bound on the information loss introduced by our compression strategy.

\subsection{Overview}
\begin{figure*}
    \centering
    \includegraphics[width=1\linewidth]{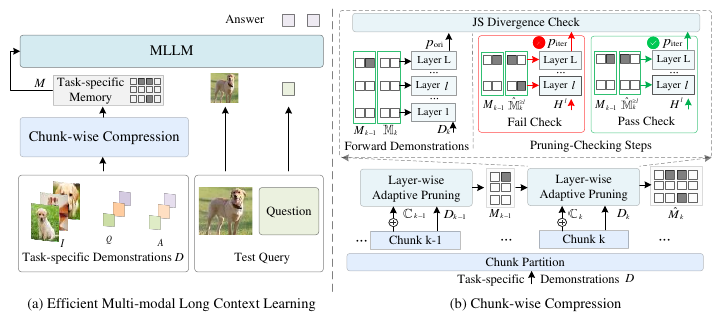}
    % \vspace{-10pt}
\caption{(a) The overall framewrk of efficient multi-modal long-context learning. (b) Chunk-wise compression with layer-adaptive pruning, where pruning steps iteratively update output probabilities and are validated using a JS divergence check. Gray squares indicate pruned tokens, with red and green arrows representing failed and successful pruning steps, respectively.
}

    \label{fig:framework}
\end{figure*}

Given a downstream multi-modal task and its demonstration examples, our goal is to adapt a pre-trained MLLM without any training or parameter fine-tuning.
To achieve this, we construct a long context $\mathbb{C}$ as the input of MLLM by concatenating $N$ task-specific demonstration examples $D$. Representing each example as $\langle {I}_i, {Q}_i, {A}_i \rangle$ consisting of an image ${I}_i$, a question $Q_i$ and a corresponding answer $A_i$, the generation of $\mathbb{C}$ can be denoted as,
\begin{equation}
 {\mathbb{C}} = \bigoplus_{i=1}^{N} \langle  {I}_i,  {Q}_i,  {A}_i \rangle.
\end{equation}
Conditioning on the multi-modal long context $\mathbb{C}$ and multi-modal test queries $X$, MLLM can generate more accurate answers $Y$ as follows:
\begin{equation}
    Y = \text{MLLM}(\mathbb{C}, X).
\end{equation}
Increasing the number of examples $N$ provides more informative contexts and enhances performance,
but significantly increases the inference cost and memory usage.

We thus propose to reduce the long context $\mathbb{C}$ into a compact task-specific memory ${M}$ through chunk-wise compression with layer-wise adaptive pruning. Using $\mathbb{M}$ to denote the key-value (KV) cache of context $\mathbb{C}$, we aim to extract a more compact memory ${M}$ from $\mathbb{M}$.

This can be achieved by dividing the $\mathbb{C}$ into multiple chunks and adaptively spotting and reserving important tokens across chunks and layers to minimize the change of output probabilities distribution. The target of ${M}$ extraction from a chunk can be conceptually formulated as
\begin{align}
{M} =  \arg & \min_{{M} \subset \mathbb{M}}|{M}|, \nonumber \\
 \text{s.t.} \quad \mathcal{D}_\mathrm{JS}(P(Y \mid \mathbb{M}, X&),  P(Y \mid {M}, X)) \leq \Delta,
 \label{eq:objective}
\end{align}
where $\mathcal{D}_\mathrm{JS}(\cdot)$ denotes the Jensen-Shannon divergence, which measures the difference between two probability distributions. $\Delta$ is a configurable upper bound regulating the trade-off between information loss and the length of $M$. 

The overall framework of EMLoC is illustrated in~\cref{fig:framework}(a). EMLoC implements the objective in Eq.~\eqref{eq:objective} in two stages. It first partitions $\mathbb{C}$ into smaller chunks to enable processing on resource-constrained devices. As shown in~\cref{fig:framework}(b), we partition $N$ demonstration examples into $K$ chunks, each containing $\frac{N}{K}$ examples. We denote each chunk as $\mathbb{C}_k$, i.e.,
\begin{equation}
    \{\mathbb{C}_k\}_{k=1:K} = \mathbb{C}.
\end{equation}
Let $D_k$ denote the individual demonstrations in chunk $\mathbb{C}_k$  before concatenation. We iteratively compress each chunk to form the compressed memory: 
\begin{equation} M_k = \mathrm{LAP}(M_{k-1}, \mathbb{C}_k, D_k), \end{equation} where $\mathrm{LAP}(\cdot)$ is our layer-wise adaptive pruning, and ${M}_{k-1}$  is the compressed memory from the previous chunks. Iterating this process from $k=1$ to $k=K$ yields a final compressed memory ${M}_K$, i.e., $M$.

EMLoC proceeds to prune the KV cache of each chunk in a layer-wise manner under a Jensen–Shannon (JS) divergence constraint, retaining only the most relevant tokens for the downstream task. It is difficult to set a global upper bound $\Delta$. An effective implementation to $\Delta$ is achieved by setting a local constraint $\delta$. The details of layer-wise compression and implementation to $\Delta$ are presented in following parts.

% \subsection{Chunk-wise Compression}
% %With a large number of examples, the multi-modal context becomes extremely long. 
% %It's expensive to directly compress the long context into a compact memory due to the prohibitive GPU memory requirements. 

% Let $D_k$ represent the demonstration examples in the $k$-th chunk. We concatenate the examples in $D_k$ to form the context $\mathbb{C}_k$ for that chunk. Each chunk is compressed sequentially, updating the memory as follows:
% \begin{equation}
% {M}_k = \mathrm{LAP}({M}_{k-1}, \mathbb{C}_k, D_k),
% \end{equation}
% where $\mathrm{LAP}(\cdot)$ is our layer-wise adaptive pruning module (detailed in \cref{sec:LAP}), and ${M}_{k-1}$ is the compressed memory after processing the first $k-1$ chunks. By iterating from $k=1$ to $k=K$, we obtain a final compressed memory ${M}_K$ that encodes all $N$ examples but at significantly reduced cost.

\subsection{Layer-wise Adaptive Pruning}
\label{sec:LAP}
% Different tasks often require different types of knowledge, leading to varying level of token redundancy. Moreover, recent studies~\cite{xiao2024efficient, cai2024pyramidkv} reveal that earlier layers in large language model tend to be more important than later ones.
% Instead of using a fixed pruning strategy that retains a constant number of tokens per layer, we propose layer-wise adaptive pruning strategy to dynamically reserve the optimal number of tokens per layer without degrading performance. The optimal number of tokens is the minimal number that ensures the Jensen-Shannon (JS) divergence between the compressed and uncompressed output distributions remains below a predefined threshold $\delta$. The detailed pseudo code is depicted in ~\cref{sup:lap}.

Different tasks require varying levels of knowledge, resulting in different degrees of token redundancy. Recent studies~\cite{xiao2024efficient, cai2024pyramidkv} show that earlier layers in large language models are more crucial than later ones. Instead of a fixed pruning strategy~\cite{li2024snapkv, cai2024pyramidkv} that retains a constant number of tokens per layer, we propose a layer-wise adaptive pruning strategy that dynamically adjusts the token number per layer while preserving performance. The optimal number of tokens per layer is the minimum required to keep the Jensen-Shannon divergence between the probability distributions conditioned on the original and compressed memory below a predefined threshold $\delta$. Detailed pseudocode is provided in~\cref{sup:lap}.
% The layer-wise adaptive pruning operates in two key phases: compressing the memory (KV cache) by adaptively pruning redundant tokens, and ensuring that the pruning does not lead to significant performance degradation as measured by the JS divergence between the probability distribution conditioned on original and compressed memory. 
The procedure proceeds as follows:

\noindent\textbf{Forward Demonstrations:} 
Given the context $ \mathbb{C}_k$ of $k$-th chunk, the KV cache $\mathbb{M}_k$ of this chunk is firstly extracted, conditioned on the previously compressed memory ${ {M}}_{k-1}$:
\begin{equation}
  \mathbb{M}_{k} = \text{ExtractKV}({ {M}}_{k-1},  \mathbb{C}_k).
\end{equation}
% where ${M}_{k-1}$ is concatenated with the ${M}_{{C}_k}$ to form the key and value in the forward of attention layer.

A forward pass is performed through the MLLM with the concatenated memory ${ {M}}_{k-1} \oplus  \mathbb{M}_{k}$ and the individual demonstration examples $D_k\in \mathbb{R}^{\frac{N}{K} \times T}$ in chunk $\mathbb{C}$. The output includes the original output probabilities $p_{\text{ori}}$ of answer tokens, attention weights $\alpha\in \mathbb{R}^{L\times (\frac{N}{K} \times T)\times S}$, and hidden states $H\in \mathbb{R}^{L\times \frac{N}{K} \times T}$:
\begin{equation}
\label{eq:forward_demonstration}
(p_{\text{ori}}, \alpha,  {H}) = \text{MLLM}({ {M}}_{k-1} \oplus  \mathbb{M}_{k},  {D}_k),
\end{equation}
where $L$ is the number of layers, $T$ denotes the length of each demonstration example, and $S$ represents the sequence length of the current chunk. Then, for each layer, we calculate the importance score of each token in $\mathbb{M}_{k}$ using the accumulated attention weights from answer tokens. The answer tokens serve as an observation window to estimate the importance of each token in chunk $\mathbb{C}$. The importance score of $j$-th token in $l$-th layer can be formulated as:
\begin{equation}
    \beta^l_j = \sum_{i\in \text{ans}\_{\text{index}}}\alpha^l_{ij},
\end{equation}
where $\text{ans}\_{\text{index}}$ is the index of answer tokens.

\noindent\textbf{Pruning-Checking Step:} 
Tokens with higher importance are retained, while less important ones are pruned iteratively from the top layer down. This top-down approach improves efficiency by avoiding forward passes through unpruned layers.  At each step, pruning is applied to a single layer using a candidate ratio $r$, greedily selected in ascending order from the retention ratio set $R$.  Once the top layers above the $l$-th layer are pruned, the pruning of the $l$-th layer is formulated as selecting the top $r\times S$ tokens based on importance scores: 
\begin{equation}
{ \hat{\mathbb{M}}}_{k}^l =  \mathbb{M}_{k}^l (\text{Topk}(\beta^l, r \times S)).
\end{equation}
As illustrated in ~\cref{fig:framework}~(b), we can get the output probabilities $p_{\text{iter}}$ of all answer tokens in one pruning step:
\begin{equation}
    p_{\text{iter}} = \text{MLLM}({M}^{\geq l}_{k-1} \oplus \hat{\mathbb{M}}^{\geq l}_{k}, H^l),
\end{equation}
where ${M}^{\geq l}_{k-1}$ denotes the compressed memory from layer $l$ onward for the previous $k$-1 chunks, $\hat{\mathbb{M}}^{\geq l}_{k}$ denotes the compressed memory from layer $l$ onward for the current $k$-th chunk, hidden states $H^l$ is the input embeddings of $l$-th layer. Leveraging hidden states $H^l$ extracted in ~\cref{eq:forward_demonstration} can reduce redundant computation in pruning steps.

After each pruning step, we perform the JS divergence check between the output distributions $p_{\text{ori}}$ and $p_{\text{iter}}$ to measure the information loss. 
If the divergence is below the threshold $\delta$, the pruning process for the current layer is successful and continues to prune the next bottom layer otherwise select a larger $r$ from rentention ratio set $R$ to prune this layer again. 
As shown in \cref{fig:framework}(b), a step marked by red arrows fails the check, while a subsequent step with green arrows retains more tokens and successfully passes the check.
This pruning-checking step ensures that the compressed memory retains the essential information for task performance while minimizing inference costs.

After successfully pruning all layer for $k$-th chunk, we can get the compressed memory ${M}_{k}$ of $k$ chunks by concatenating ${M}_{k-1}$ and $\hat{\mathbb{M}_{k}}$.

\subsection{Theoretical Analysis of Information Loss}
\label{sec:upper_bound}
It is difficult to set a global upper bound $\Delta$ for JS divergence when sequentially processing multiple chunks. An effective implementation to $\Delta$ is achieved by setting a local constraint $\delta$ at each pruning-checking step. In this part, we theoretically analyze how the local constraint $\delta$ contributes to the global JS divergence $\Delta$ between the full memory $\mathbb{M}$ and its compressed counterpart $M$. 

We assume that the $N$ demonstration examples $D$ follow the same data distribution as the test set. Thus, during inference with the memory $M$ , the probability distribution $P_M$ of test set equals to that of demonstration examples $D$:
\begin{equation}
    P_{M} = P({Y}|M, {X}) = P(\pi_{A}(D)|M, \pi_{I,Q}(D))=P_{M}^{D},
\end{equation}
% \begin{equation}
%     P_{M}^{D} = P(Y|M, X) = P(D_{Y}|M, D_{X}),
%     % P_{M} = P(D_{Y}|M, D_{X}),
% \end{equation}
where $\pi_{I,Q}$ denotes extracting the images and questions as multi-modal questions and $\pi_{A}$ denotes getting the corresponding answers in the demonstration set.

\noindent\textbf{Local JS Distance Constraint:}With the individual demonstrations $D_k$ of the $k$-th chunk, the JS distance between the probability distributions under ${M}_{k-1} \oplus \mathbb{M}_{k}$ and ${M}_k$ is bounded by a divergence distance threshold $\sqrt{\delta}$:
\begin{equation}
\label{eq:local_constraint}
    \mathcal{D}_\mathrm{JS}\left(P_{{M}_{k-1} \oplus \mathbb{M}_{k}}^{D_k}, P_{{M}_k}^{ D_k}\right) \leq \sqrt{\delta},
\end{equation}
where $\mathcal{D}_\mathrm{JS}(\cdot, \cdot)$ denotes JS distance, the arithmetic square root of JS divergence.

\noindent\textbf{Bounding the Global JS Distance:}
% By the triangle inequality for JS distance, the total JS distance between the full memory and the compressed memory of 
% $k$ chunks satisfies:
To extend the local constraint to the global upper bound, we use the triangle inequality for the JS distance:
\begin{align}
\label{eq:triangle_inequal}
    \mathcal{D}_\mathrm{JS}\left(P_{\mathbb{M}}^D, P_{{M}_k}^{D}\right) \leq &\mathcal{D}_\mathrm{JS}\left(P_{\mathbb{M}}^D, P_{{M}_{k-1}\oplus \mathbb{M}_{k}}^D\right) + \nonumber \\
    & \mathcal{D}_\mathrm{JS}\left(P_{{M}_{k-1}\oplus \mathbb{M}_{k}}^D, P_{{M}_k}^D\right)
\end{align}

The memory $M_{k-1}$ is a subset of $M_{k-1} \oplus \mathbb{M}_{k}$, which, in turn, is a subset of $\mathbb{M}$. Thus, $M_{k-1} \oplus \mathbb{M}_{k}$ more closely resembles the original full memory $\mathbb{M}$. Consequently, the probability distribution $P_{M_{k-1} \oplus \mathbb{M}_{k}}^D$, conditioned on $M_{k-1} \oplus \mathbb{M}_{k}$, is more similar to $P_{\mathbb{M}}^D$ than to $P_{M_{k-1}}^D$:

\begin{equation}
\label{eq:recursive}
    \mathcal{D}_\mathrm{JS}\left(P_{\mathbb{M}}^D, P_{{M}_{k-1}\oplus \mathbb{M}_{k}}^D\right) \leq \mathcal{D}_\mathrm{JS}\left(P_{\mathbb{M}}^D, P_{{M}_{k-1}}^D\right).
\end{equation}
In other words, adding more chunks from the full memory reduces the divergence with $P^D_{\mathbb{M}}$.
Moreover, because the demonstrations $D_k$ are most closely related to the context in the $k$-th chunk itself, we empirically assume:
\begin{align}
\label{eq:full_to_local}
    \mathcal{D}_\mathrm{JS}\left(P_{{M}_{k-1}\oplus \mathbb{M}_{k}}^D, P_{{M}_k}^D\right) \leq \mathcal{D}_\mathrm{JS}\left(P_{{M}_{k-1}\oplus \mathbb{M}_{k}}^{D_k}, P_{{M}_k}^{D_k}\right)
\end{align}
% Besides, when compressing $k$-th chunk, the demonstrations $D_k$ of current chunk is strongly related to the context. It demonstrates that the compression of $k$-th chunk have a more impact on the distribution of subset $D_k$ than the overall demonstration set $D$. As a result, we conclude the empirical assumption that:

\noindent\textbf{Theoretical Upper Bound:}
Combining the \cref{eq:local_constraint,eq:triangle_inequal,eq:recursive,eq:full_to_local}, we can derive the following global upper bound of the information loss:
\begin{equation}
\label{eq:final_up_bound}
    \mathcal{D}_\mathrm{JS}\left(P_{\mathbb{M}}^{D}, P_{{M}_K}^D\right) \leq
    (K-1)\sqrt{\delta} + \epsilon, 
\end{equation}
where the $\epsilon$ is the JS distance between the probability distribution of full context $\mathbb{C}$ and that of the first uncompressed chunk $\mathbb{C}_1$.
Hence, by applying the local constraint at each chunk-compression step, the global JS distance $\sqrt{\Delta}$ between the compressed memory and the full memory is controlled by $\delta$ and chunk number $K$. More proof and analysis are depicted in ~\cref{sup:linear_upper_bound}.

\section{Experiments}

\begin{table*}[ht]
\small
\centering
\caption{Results of efficient multi-modal long context learning (EMLoC) in various downstream tasks. The value in the gray cell is the context length.  $\dagger$ denotes using 50 examples. $\ddagger$ represents utilizing 200 examples.  }
\label{tab:tab1_main}
\begin{tabular}{c|c|cccccc}
\toprule
Method              & \makecell{Example\\Number} & ImageNet100 & ScreenSpot & MME-RW & IllusionVQA & OK-VQA & YouCook2  \\ \hline
Llava1.5~(7B)          &   \multirow{5}{*}{0 }  &       12.3     &       9.7     &   28.2     &      24.1       &   53.6     &     -   \\
InternVL2~(8B)         &     &      12.5       &      2.7      &     33.9   &       28.0      &   47.1     &     88.0     \\
% InternVL2.5~(8B)         &     &     30.2        &      10.2      &    33.7   &      34.4      &   54.0     &    110.9     \\
Llama3.2-V~(11B)          &    &      47.6       &      8.1      &  14.6      &  33.0           &   -     &    -      \\
MiniCPM-V2.6~(8B)       &    &      31.0       &      0.3      &     37.2   &      34.6       &    48.3    &   3.3       \\
Qwen2-VL~(7B)            &    &   28.0          &      14.2      &    36.6    &     35.3        &    52.1    &       25.4   \\ \hline
% MiniCPM-V 2.6       & 10    &             &            &        &             &        &          \\
\multirow{4}{*}{\makecell{MLoC\\(Qwen2-VL) }}           & \multirow{2}{*}{5}   &     43.2 $\dagger$      &     14.7      &    39.4   &   38.8        &     58.4   &     86.9     \\ 
 &    & \cellcolor{lightgray}4109 & \cellcolor{lightgray}1996  & \cellcolor{lightgray}1924 &  \cellcolor{lightgray}1826 & \cellcolor{lightgray}1401  & \cellcolor{lightgray}5907  \\ \cline{2-8}
      & \multirow{2}{*}{20}    &         62.6$\ddagger$    &      18.2      &     41.1   &     \textbf{40.9}        &     58.6   &    
\textbf{108.8}\\ 
&    & \cellcolor{lightgray}16264 & \cellcolor{lightgray}7905  & \cellcolor{lightgray}7393 & \cellcolor{lightgray}7271  &   \cellcolor{lightgray}5730 & \cellcolor{lightgray}23464  \\ \hline 

% \multirow{2}{*}{Qwen2-VL }           & 5    &     43.6        &     14.7      &    39.4   &   38.8         &     57.9   &     86.9     \\ 
% Qwen2-VL      & 20    &         62.62     &      18.2      &     41.1   &     40.9        &     57.09   &     105.1(10 shot)   
% 108.8(20 shot)\\ 
\multirow{2}{*}{\makecell{EMLoC\\(Qwen2-VL)}
}     & \multirow{2}{*}{20}    &     \textbf{63.6} $\ddagger$     &     \textbf{18.3}   &   \textbf{42.2}    &    \textbf{40.9}      &   \textbf{58.7}  &  {102.0}   \\ 
&    & \cellcolor{lightgray}3643 & \cellcolor{lightgray}1415  & \cellcolor{lightgray}1510 &  \cellcolor{lightgray}1878 &  \cellcolor{lightgray}934 & \cellcolor{lightgray}6218  \\

\bottomrule
\end{tabular}
\vspace{-2pt}
\end{table*}

\subsection{Experimental Setting}
\noindent\textbf{Evaluation Dataset}.
 We evaluate our EMLoC on six challenging benchmarks: \textit{ImageNet100}, a subset of ImageNet-1k~\cite{deng2009imagenet} with the first 100 classes for recognition, \textit{ScreenSpot} for cross-platform GUI grounding, 
\textit{MME-RW} for real-world multimodal tasks, 
\textit{IllusionVQA} for illusion understanding, 
\textit{OK-VQA} for knowledge-based QA, and 
\textit{YouCook2} for video understanding. For datasets without predefined validation splits, we randomly sample 100 test examples for evaluation.

\noindent\textbf{Implementation Details.} Experiments are conducted on NVIDIA L20 GPUs with 48GB of memory. Inference time is measured with a batch size of 1 on one GPU. The default JS divergence threshold $\delta$ is set to 0.005, and the chunk size is 1.6k. The retention ratio set $R$ is [0.1, 0.2, 0.5, 1.0]. We use Qwen2-VL as the base model due to its support for 32k multi-modal long contexts in any format and a vision encoder that processes images at any resolution without tiling. For the ImageNet100 dataset, the image resolution is 224×224, while other benchmarks use 448×448. The evaluation of other MLLMs uses their default image resolution.

\subsection{Experimental Results}
\noindent\textbf{Performance of EMLoC on multiple tasks.}
% Table~\ref{tab:tab1_main} compares standard zero-shot inference in MLLMs with vanilla multi-modal long-context learning (MLoC) and our proposed EMLoC.
As shown in \cref{tab:tab1_main}, multi-modal long-context learning significantly enhances Qwen2-VL's performance across all downstream tasks, whether using 5 or 20 demonstration examples. In tasks like ImageNet100 and YouCook2, leveraging an extremely long context yields substantial performance gains.

These results underscore the importance of efficient multi-modal long context learning for practical applications, particularly in ImageNet100 and YouCook2. When utilizing 20 examples, our EMLoC dramatically reduces the average context length from \textit{11338} to \textit{2600}, a remarkable 77\% reduction, without sacrificing performance. Notably, our EMLoC outperforms MLoC with 5 demonstration examples and \textit{3530} average context length by a wide margin, even with a shorter context length.
Furthermore, when utilizing 20 examples, EMLoC surpasses MLoC with full memory in most benchmarks. 
This improvement is likely due to the removal of irrelevant background noise or distractions during compression. As a result, the compressed memory is more precise, leading to better performance.

Meanwhile, inference costs have been significantly reduced. On ImageNet100 with 200 examples, EMLoC reduces \emph{inference FLOPs} from 1.76T to 1.35T and total \emph{inference time} from 1866s to 1107s compared to MLoC.

\noindent\textbf{Results with varying numbers of examples.} 
~\cref{tab:tab3} shows how the performance of MLoC and EMLoC changes as the number of examples increases on the ImageNet100 dataset. Both MLoC and EMLoC show steady improvement with more examples, demonstrating strong long-context learning capabilities. However, this improvement comes with a sharp rise in context length, which inflates computational overhead. In contrast, EMLoC compresses the context by nearly a quarter while maintaining or even surpassing MLoC’s accuracy (e.g., 63.7 vs. 62.6 with 200 examples). This significant reduction in context length substantially decreases inference time while preserving performance.
\begin{table}[t]
\small
\centering
\caption{Multi-modal long context learning with varying numbers of examples on ImageNet 100. The value in the gray cell is the context length.}
\label{tab:tab3}
\begin{tabular}{c|ccccc}
\toprule
\multicolumn{1}{c|}{\multirow{2}{*}{Method}} & \multicolumn{5}{c}{Number of Examples} \\ \cline{2-6} 
\multicolumn{1}{c|}{}        & 0   & 25   & 50    & 200 & 300  \\ \midrule
\multirow{2}{*}{MLoC}                                &   28.0  &  39.9   &  43.2    &    62.6   & 62.9  \\
    &  \cellcolor{lightgray}0   & \cellcolor{lightgray}2053   & \cellcolor{lightgray}4109     &  \cellcolor{lightgray}16264 & {\cellcolor{lightgray}24468 }     \\
\hline
\multirow{2}{*}{EMLoC}                                    &   28.0  &   39.4   &  46.2    &  63.7  & 62.6 \\ 
&  \cellcolor{lightgray}0   &  \cellcolor{lightgray}565   &  \cellcolor{lightgray}946   & \cellcolor{lightgray}3643 & \cellcolor{lightgray}5365   \\
\bottomrule
\end{tabular}
\vspace{-2pt}
\end{table}

\noindent\textbf{Comparison with other multi-modal ICL methods.}
We have compared EMLoC with two other multi-modal in-context learning methods, RICES~\cite{alayrac2022flamingo} in Flamingo and MTV~\cite{huang2024multimodal}, on ImageNet100, MME-RW, and OK-VQA in Table~\ref{tab:mmicl_comparison}. RICES retrieves the top 25\% most relevant in-context samples from all samples. MTV extracts the mean activation of in-context examples as task vectors and finds the optimal replacement position of these task vectors. During inference, MTV replaces these task vectors at the optimal position of the test sample, which fails to facilitate these tasks. Our {EMLoC achieves better average performance} across the three benchmarks. It's worth noting that RICES is an online retrieval-augmented method, so it needs to forward the retrieved long context during each inference step. \emph{RICES takes 5 hours inference time and 43G memory cost on ImageNet100, while our EMLoC requires only 18 minutes with 18G memory}, showing clear advantages in efficiency.

\begin{table}[t]
\label{tab:mmicl_comparison}
\small
\centering
\caption{Comparison with other multi-modal in-context learning (ICL) methods.}
\begin{tabular}{c|ccc}
\toprule
Method & ImageNet100 & MME-RW & OK-VQA  \\
\hline
MTV    & 32.7        & 27.8   & -     \\
RICES  & \textbf{64.5} & 40.5   & 58.5   \\
\midrule
EMLoC  & 63.7        & \textbf{42.2} & \textbf{58.7} \\
\bottomrule
\end{tabular}
\end{table}

\begin{table}[t]
\centering
\small
\setlength{\tabcolsep}{0.8mm}
\caption{Comparison with fine-tuning methods on ImageNet100 with 200 examples, MME-RW with 20 examples, and OK-VQA with 20 examples.}
\label{tab:tab_peft_lora_ff}
\begin{tabular}{c|cccc}
\toprule
Method & ImageNet100 & MME-RW & OK-VQA & Average \\
\midrule
M$^2$PT & \textbf{65.2} & 31.6 & 15.6 & 37.5 \\
VPT   & 43.6 & 38.7 & 54.5 & 45.6 \\
% VPT*  & 61.2 & 35.1 & 34.8 \\
E$^2$PT & 48.6 & 39.0 & 55.8 & 47.8 \\
Full Fine-tuning & 64.7 & \textbf{42.7} & 49.7 & 52.4 \\
LoRA             & 61.1 & 42.1 & \textbf{60.9} & 54.7 \\
\hline
EMLoC            & 63.7 & 42.2 & 58.7 & \textbf{54.9} \\
\bottomrule
\end{tabular}
\end{table}

\noindent\textbf{Comparison with fine-tuning methods.} 
EMLoC is a training-free adaptation method that does not update any parameters of the MLLM. With less adaptation time (144s), EMLoC is comparable to LoRA (234s)~\cite{hu2022lora} and full fine-tuning (820s). Our EMLoC also surpasses other PEFT methods, such as M$^2$PT~\cite{wang2024m2pt}, VPT~\cite{jia2022visual}, and E$^2$PT~\cite{han2023e2vpt}. The training details of these PEFT methods and full fine-tuning are depicted in ~\cref{sup:imple_details}.

\subsection{Ablation Studies}
\label{sec:exp_ablation}
\textbf{Comparison with other pruning strategies.}
% Our layer-wise adaptive pruning strategy dynamically assesses the importance of various layers and tokens using the JS divergence of the output distribution. Compared to fixed pruning strategies, our EMLoC approach is able to retain more critical tokens under the same compression ratio, thereby delivering superior performance. As shown in ~\cref{tab:tab4_pruning_strategy}, on ImageNet100, our method sustains performance at a compression ratio of 22.4\%, whereas the performance of other pruning methods declines significantly. However, at a compression ratio of 15.7\% with $\delta$=0.02, our method's performance drops to 57.7\%, indicating that some essential tokens identified by our strategy were pruned at this level of compression. In contrast, other methods do not exhibit significant performance drops, suggesting that their pruning strategies failed to effectively identify critical tokens at the compression ratio of 22.4\%.

The layer-adaptive pruning strategy employs JS divergence analysis to dynamically evaluate token importance across layers. Unlike fixed pruning approaches, EMLoC maintains superior performance through selective token preservation. As evidenced in ~\cref{tab:tab4_pruning_strategy}, at 22.4\% compression on ImageNet100, EMLoC sustains 63.7 while competitors suffer significant degradation. This demonstrates our method's effectiveness in identifying and retaining critical tokens.
The performance boundary emerges at extreme compression (15.7\%, $\delta$=0.02), where EMLoC's accuracy drops to 57.7, indicating essential token removal, while conventional methods show paradoxical stability~(47.3-48.2). This suggests competing approaches fail to properly distinguish crucial tokens even at moderate compression (22.4\%), as their performance collapses before reaching critical pruning thresholds. A more comprehensive comparison with additional pruning strategies across various metrics can be found in Appendix~\ref{sup: pruning_strategy}.

\begin{table}[t]
\centering
\small
\caption{Comparison with other pruning strategies under the different compression ratios on ImageNet100 with 200 demonstrations and MME-RW with 20 demonstrations}
\label{tab:tab4_pruning_strategy}
\begin{tabular}{c|cc|cc}
\toprule
\multicolumn{1}{c|}{\multirow{2}{*}{Method}} &\multicolumn{2}{c|}{ImageNet100} &  \multicolumn{2}{c}{MME-RW} \\ \cline{2-5}

& Ratio & Acc & Ratio & Acc \\ \midrule
 MLCL &100\% & 62.6  & 100\% & 41.1 \\ \hline
 SnapKV/H2O &  22.4\% & 47.6 & 20.4\% & 39.6 \\
 PyramidKV &  22.4\% & 49.3 & 20.4\% & 39.8 \\
 EMLoC(Ours) &  22.4\% & \textbf{63.7} & 20.4\% & \textbf{42.2}\\ \hline
SnapKV/H2O &  15.7\% & 47.3 & 14.1\% & 40.2 \\
 PyramidKV &  15.7\% & 48.2 & 14.1\% & 39.4\\
 EMLoC(Ours) &  15.7\% & \textbf{57.7} & 14.1\% & \textbf{40.6}
 \\  \bottomrule
\end{tabular}
\vspace{-2pt}
\end{table}

\textbf{Results of different $\delta$.}
\begin{figure}[t]
    \centering
    \includegraphics[width=1\linewidth]{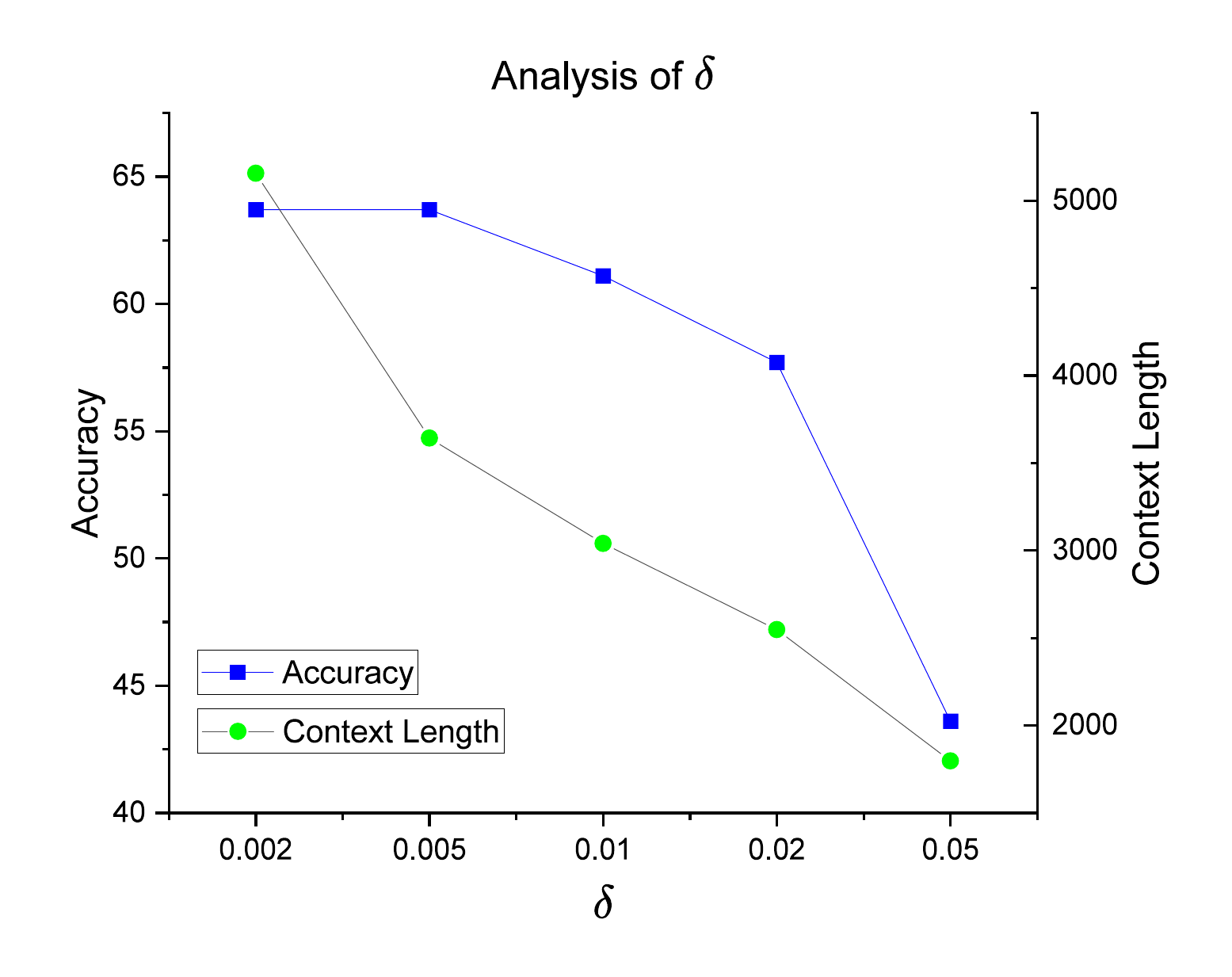}
    \vspace{-2pt}
    \caption{Performance and context length trends of EMLoC on ImageNet100 with 200 examples across different $\delta$ values}
    \label{fig:delta}
    \vspace{-2pt}
\end{figure}
\cref{fig:delta} presents the effect of varying the Jensen-Shannon divergence threshold. A higher threshold results in greater compression but also leads to increased information loss. As the threshold increases, both accuracy and context length decrease. It demonstrates that $\delta$ can effectively control the global upper bound $\Delta$. The default threshold of 0.005 is selected as a balance between compression efficiency and performance preservation.

\textbf{Results in various chunk lengths.}
\cref{tab:tab5_chunk_number} examines chunk length impacts on computational efficiency and compression effectiveness. Longer chunks linearly increase memory demands and compression ratios while maintaining stable adaptation times (150s) and accuracy (62.4-63.7). Our chunk-wise compression enables memory-efficient processing without performance degradation - notably supporting consumer GPUs (NVIDIA 4090) at 0.8k chunks with merely 1.3 accuracy drop. The default 1.6k configuration balances memory usage (38G) and compression effectiveness (22.4 ratio), while extreme 3.2k lengths cause OOM errors.

\begin{table}[t]
\centering
\small
\caption{Results in various chunk lengths and chunk numbers on ImageNet100 with 200 demonstrations.}
\label{tab:tab5_chunk_number}
\begin{tabular}{c|ccccc}
\toprule
\makecell{Chunk\\Length} & \makecell{Chunk\\Number}& \makecell{Adaptation\\Time} & Acc & \makecell{Peak\\Memory} & Ratio \\ \midrule
     0.8k   &20 &     150s &  62.4 &   24G  &      20.4                                                         \\
     1.6k  &   10  &    144s   &  \textbf{63.7} &  38G   &    22.4                                                                                                         \\
       % 2.0k    &8&     145s  &  62.2 &   42G  &    23.1                                                                                                        \\
      2.4k     &7 &    149s   &  63.4   &       44G                                                &      24.7                                                  \\
      2.8k     &6 &    142s   &  62.5   &       48G                                                &      26.3                                                 \\
    3.2k     & 5 &  -     &  -   &       OOM       &      -                                                 \\
    
       \bottomrule
\end{tabular}
\vspace{-2pt}
\end{table}

\textbf{Results with different retention ratios.} 
The retention ratio $r$ is selected greedily from an ascending predefined retention ratios set $R$, progressing to larger ratios until satisfying the JS divergence constraint or reaching 1.0. Two distinct patterns emerge based on the minimum ratio setting:
When significantly below optimal compression levels (e.g., 0.05 vs 22.4\% optimal), 
$\delta$ becomes the dominant control factor. As shown in Table 6 and Figure 3, this configuration enables adaptive balancing - excessive layer compression is offset by others retaining more tokens to meet JS constraints, maintaining comparable overall ratios (21.0\% vs 22.4\%).
Near-optimal minimum ratios (e.g., 0.2) directly determine final compression outcomes (30.4\%).
Based on empirical observations, we recommend setting the minimum ratio to approximately half of the optimal compression ratio. If the minimum ratio is too small, the overall compression ratio may not be sufficiently reduced, while simultaneously discarding both important and redundant tokens at some layers. The final retention ratios $R$ are set to [0.1, 0.2, 0.5, 1.0].
% which provides effective granularity for constraint satisfaction while maintaining compression efficiency.

\begin{table}[t]
\centering
\small
\caption{Results under different retention ratios.}
\label{tab:tab6_retention_ratio}
\begin{tabular}{c|ccc}
\toprule
\makecell{Retention\\Ratios} & \makecell{Context\\Length} & Ratio & Acc  \\ \midrule
 0.1, 0.2, 0.5, 1.0 & 3643& 22.4\% & \textbf{63.7} \\
 0.2, 0.5, 1.0 & 4952 & 30.4\% & 61.6 \\ 
 0.05, 0.1, 0.2, 0.5, 1.0 & 3421 & 21.0\% & 58.6 \\
  0.1, 0.2, 0.5, 0.8, 1.0 & 3728 & 22.9\% & 62.2 \\
 % 0.1, 0.5, 1.0 & 3441 & \% & 62.6 \\  
  \bottomrule
\end{tabular}
\end{table}

\textbf{Abalation studies observation window.}
For each token in the multi-modal long context, we use the answer tokens as the observation window to compute its importance score. The answer tokens are the most critical elements in the Image-Question-Answer pair, as they encapsulate the key semantic information of the example. 
In ~\cref{tab:tab7_observation}, we compare different observation windows. The results show that answer tokens serve as an effective observation window, while the question tokens and image tokens often include a significant amount of irrelevant or noisy data. Furthermore, as highlighted in the last row of ~\cref{tab:tab7_observation}, retaining answer tokens yields better performance, further confirming that answer tokens are crucial in multi-modal contexts.
\begin{table}[t]
\centering
\small
\caption{Abalation studies observation window. $*$ indicates preserving all answer tokens in the memory.}
\label{tab:tab7_observation}
\begin{tabular}{c|ccc}
\toprule
\makecell{Observation\\Window} & \makecell{Context\\Length} & Ratio & Acc  \\ \midrule
 Answer & 3643 & 22.4\% & 61.9 \\
 Question+Answer & 5304 & 32.6\% & 58.3 \\
 Image+Question+Answer & 8796 & 54.0\% & 55.4 \\ 
 Answer$*$ & 3643 & 22.4\% & \textbf{63.7} \\
 % 0.1, 0.5, 1.0 & 3441 & \% & 62.6 \\  
  \bottomrule
\end{tabular}
\end{table}

% \textbf{Extension to More number shots}

\subsection{Visualization of Compression}
\noindent\textbf{Remaining token number across layers}
Previous studies~\cite{xiao2024efficient,cai2024pyramidkv} suggest that earlier layers in large language models (LLMs) are more important than later ones, advocating for a pyramid-shaped pruning strategy. However, our layer-wise adaptive pruning approach challenges this assumption, arguing that layer importance should be determined dynamically based on task-specific demonstrations rather than a fixed heuristic.

\begin{figure}[t]
    \centering
    \includegraphics[width=1\linewidth]{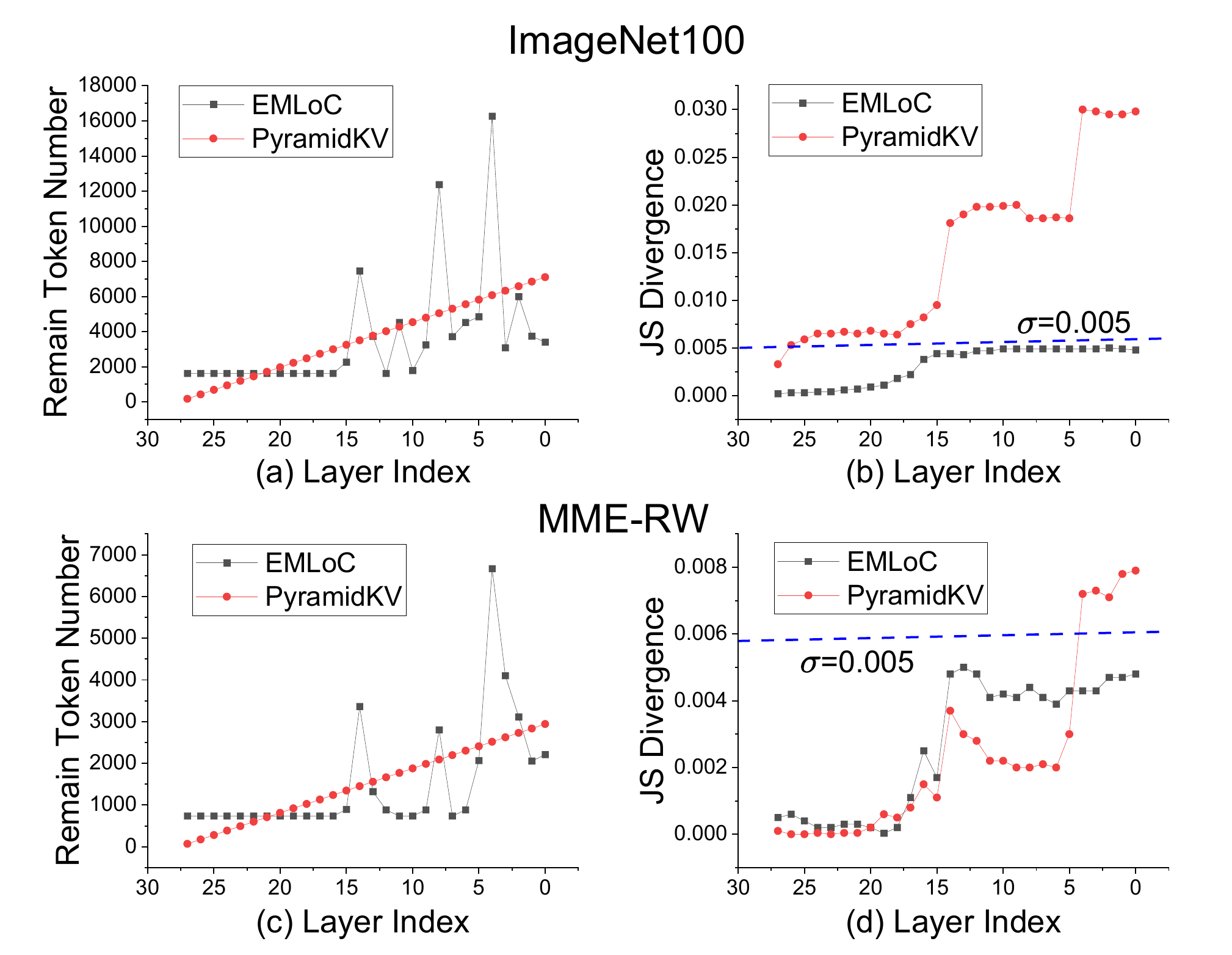}
    \vspace{-10pt}
    \caption{Remaining token number of  EMLoC and PyramidKV in ImageNet100 with 200 demonstrations and MME-RW with 20 demonstrations. The corresponding JS divergence after pruning is also illustrated to demonstrate the advantage of EMLoC.}
    \label{fig:index_visualization}
    \vspace{-10pt}
    
\end{figure}

As shown in \cref{fig:index_visualization}, experiments on ImageNet100 and MME-RW reveal that layer importance does not strictly follow a pyramid structure. Instead, certain early layers are particularly critical, as indicated by the spikes in \cref{fig:index_visualization}(a) and (c). Pruning tokens in these key layers causes significant shifts in the model’s output distribution. For instance, in \cref{fig:index_visualization}(b) and (d), applying PyramidKV pruning to these layers leads to a sharp increase in JS divergence, significantly degrading performance. The corresponding performance metrics are provided in \cref{tab:tab4_pruning_strategy}.
Our findings highlight that layers 4, 8, and 14 of Qwen2-VL are crucial in both ImageNet100 and MME-RW, as they retain significantly more tokens than adjacent layers.

\begin{figure}[t]
    \centering
    \includegraphics[width=0.9\linewidth]{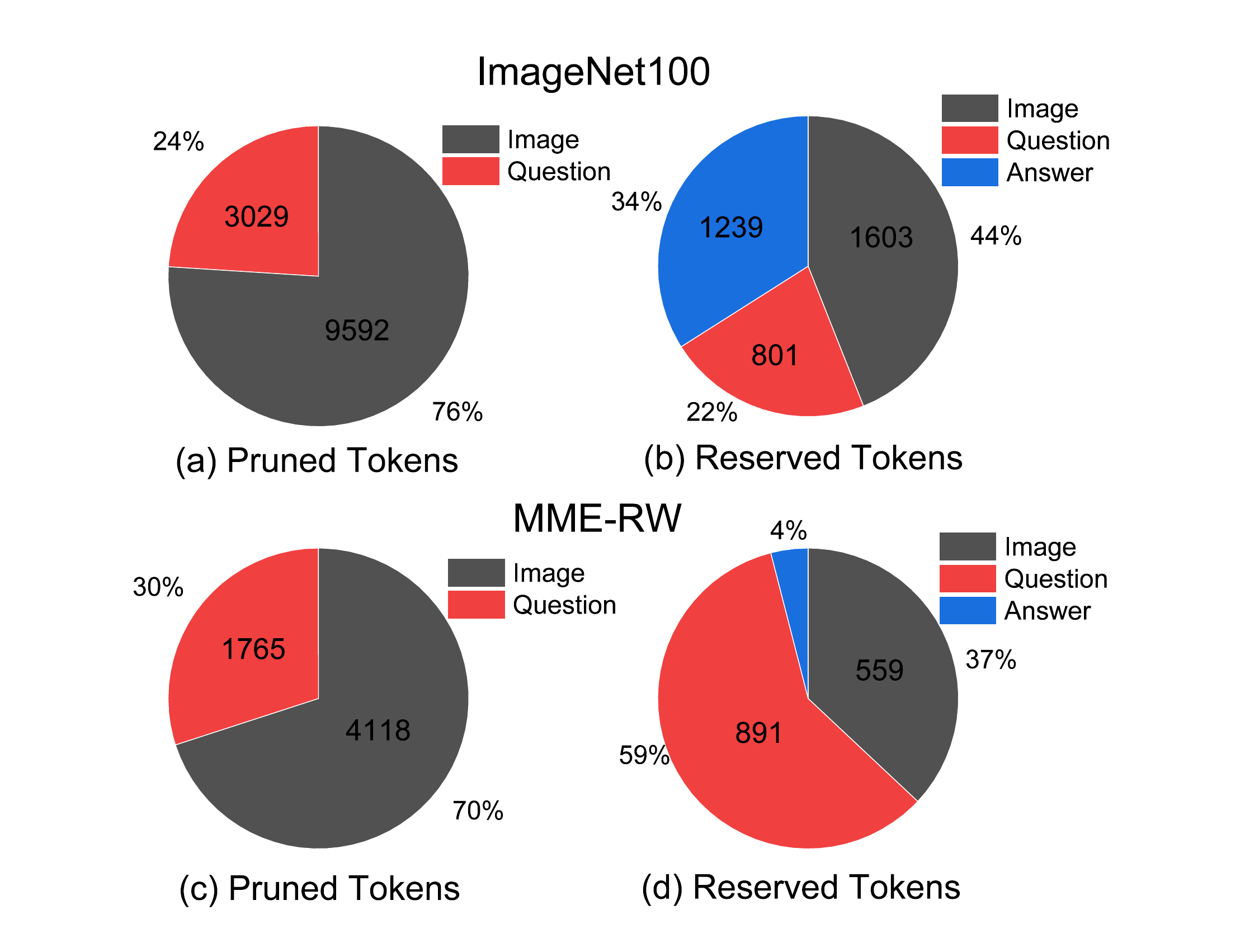}
    \vspace{-10pt}
    \caption{Distribution of pruned and reserved tokens.}
    \label{fig:distribution_visualization}
    \vspace{-10pt}
\end{figure}

\noindent\textbf{Distribution of Pruned and Reserved Tokens.}
This study serves as an initial exploration of multi-modal context compression. As shown in \cref{fig:distribution_visualization}(a) and (c), image tokens make up the majority of pruned tokens, suggesting that the visual modality contains more redundancy than the textual modality. Besides, \cref{fig:distribution_visualization}(b) and (d) indicate that fewer image tokens are retained compared to text tokens, demonstrating effective pruning of redundant visual information. Additionally, variations in tasks may lead to differences in their compression rates.

% \section{Conclusion} This paper introduces Efficient Multi-modal Long Context Learning(EMLoC) for training-free adaptation. EMLoC leverages chunk-wise compression with layer-wise adaptive pruning to compress multi-modal long contexts into a compact memory. Experiment results show that EMLoC significantly reduces inference overhead while preserving performance on various vision-language tasks. Future research may explore more fine-grained importance metrics and extend EMLoC to other multi-modal scenarios. 

\section{Conclusion} This paper presents EMLoC, a training-free method combining chunk-wise compression with layer-wise adaptive pruning to build a compact, task-specific memory for downstream tasks. Experiment results show EMLoC reduces inference overhead while preserving strong performance in multiple vision-language tasks, providing a scalable solution for efficient multi-modal long context learning.

\section*{Acknowledgements}
This work is supported in part by Grant No. 2023-JCJQ-LA-001-088, in part by the Natural Science Foundation of China under Grant No. U20B2052, 61936011, 62236006, in part by the Okawa Foundation Research Award, in part by the Ant Group Research Fund, and in part by the Kunpeng\&Ascend Center of Excellence, Peking University.

\section*{Impact Statement}
This paper presents work whose goal is to advance the field of 
Machine Learning. There are many potential societal consequences 
of our work, none which we feel must be specifically highlighted here.
\nocite{langley00}

\bibliography{example_paper}
\bibliographystyle{icml2025}

%%%%%%%%%%%%%%%%%%%%%%%%%%%%%%%%%%%%%%%%%%%%%%%%%%%%%%%%%%%%%%%%%%%%%%%%%%%%%%%
%%%%%%%%%%%%%%%%%%%%%%%%%%%%%%%%%%%%%%%%%%%%%%%%%%%%%%%%%%%%%%%%%%%%%%%%%%%%%%%
% APPENDIX
%%%%%%%%%%%%%%%%%%%%%%%%%%%%%%%%%%%%%%%%%%%%%%%%%%%%%%%%%%%%%%%%%%%%%%%%%%%%%%%
%%%%%%%%%%%%%%%%%%%%%%%%%%%%%%%%%%%%%%%%%%%%%%%%%%%%%%%%%%%%%%%%%%%%%%%%%%%%%%%
\newpage
\appendix
\onecolumn
\section{Pseudo Code of Layer-wise Adaptive Pruning}
\label{sup:lap}
\begin{algorithm}[H]
\caption{Layer-wise Adaptive Pruning}
\label{alg:lap}

\KwIn{
    $\mathbb{C}_k\in \mathbb{R}^{S}$: concatenated context of $k$-th chunk, where $S=\frac{N}{K} \times T$;\\
    $D_k\in \mathbb{R}^{\frac{N}{K} \times T}$: demonstration examples of $k$-th chunk in a batch;\\
    ${M}_{k-1}$: KV cache of previous $k-1$ chunks;\\
    
}
\KwOut{
    ${M}_{k}$: Compressed memory of $k$ chunks.
}

$\mathbb{M}_k \gets \text{ExtractKV}({M}_{k-1}, \mathbb{C}_k)$\;

$p_{\text{ori}}, \alpha, H \gets \text{MLLM}({M}_{k-1} \oplus \mathbb{M}_{k}, D_k)$\;

$\hat{\mathbb{M}}_{k} \gets \mathbb{M}_{k}$\;

\For{$l \gets L$ \textbf{to} $1$}{
    $\beta^l \gets \left\{\sum_{i \in \text{ans\_index}} \alpha^l_{ij} \right\}_{j=1}^S$\;
 
    \For{$r \in R$ \tcp*{$R$ is retention ratio set in ascending order}}{
        $\hat{\mathbb{M}}_{k}^l \gets \mathbb{M}_{k}^l \left(\text{Topk}(\beta^l, r \times S)\right)$\; \tcp*{$S$ is chunk length}
        
        $p_{\text{iter}} \gets \text{MLLM}({M}^{\geq l}_{k-1} \oplus \hat{\mathbb{M}}^{\geq l}_{k}, H^l)$\;
        
        $\ell \gets \text{JS}(p_{\text{ori}}, p_{\text{iter}})$\;
        
        \If{$\ell \leq \delta$}{ 
            \textbf{break}\; \tcp*{$\delta$: JS divergence threshold}
        }
    }
}

${M}_{k} \gets {M}_{k-1} \oplus \hat{\mathbb{M}}_{k}$\;

\end{algorithm}

\section{Prompt Template}
In ImageNet100, 200 multi-modal examples are evenly divided into 10 chunks. Each image (224$\times$224) is encoded into approximately 64 tokens (may vary slightly due to dynamic aspect ratios), and the corresponding question-answer pair adds around 20 tokens, resulting in about 80 tokens per example. Each chunk contains 20 examples (roughly 1.6k tokens). The system prompt appears only at the start of the first chunk. Below is an example structure:

\colorbox{gray!10}{%
  \parbox{\dimexpr\linewidth-2\fboxsep}{
\ttfamily
\noindent
\# Start of 1st chunk\\
<|im\_start|> system\textbackslash n You are a helpful assistant.<|im\_end|>\\
\#\# sample 1\\
<|im\_start|> user\textbackslash n <|vision\_start|> <Image1.jpg> <|vision\_end|> What category does the image belong to? <|im\_end|>\\
<|im\_start|> assistant\textbackslash n <class 1>. <|im\_end|> \ldots\\
\# Start of 2nd chunk\\
\#\# sample 21\\
<|im\_start|> user\textbackslash n <|vision\_start|> <Image21.jpg> <|vision\_end|> What category does the image belong to? <|im\_end|>\\
<|im\_start|> assistant\textbackslash n <class 11>. <|im\_end|> \ldots
  }
}

For other image benchmarks, each image is encoded into 256 tokens (448$\times$448 resolution). Each chunk has 4 examples, resulting in a chunk size of 1.1k--1.6k tokens. For the YouCook2 video benchmark, each video with 8 frames is encoded into 1024 tokens, with 4 videos per chunk, yielding a 4.7k chunk size. If sample lengths vary significantly, we use a greedy algorithm to progressively fill each chunk up to a maximum size.

\section{More Experiments}
\subsection{Comparison with other Pruning Strategies}
\label{sup: pruning_strategy}

\begin{table}[h]
\small
\centering
\caption{Comparison with other context compression methods on ImageNet100. EMLoC* increase the chunk number from 10 to 20 and utilize a group-wise strategy to save adaptation memory and time.}
\label{tab:more_prune_strategy}
\begin{tabular}{c|cccccc}
\toprule
\textbf{Method} & \textbf{Retention Ratio} & \textbf{Adapt Time} & \textbf{Adapt Memory} & \textbf{Infer Time} & \textbf{Infer Memory} & \textbf{Acc} \\
\midrule
MLoC          & 100\%         & 28s   & 62G  & 31m  & 19G  & 62.6 \\
PyramidKV     & 22.4\%        & 54s   & 34G  & 19m  & 17G  & 49.3 \\
FastGen       & 36.0\%        & 45s   & 38G  & 37m  & 21G  & 49.3 \\
PyramidInfer  & 24.6\%        & 41s   & 42G  & 21m  & 17G  & 55.6 \\
{EMLoC}    & \textbf{22.4\%} & 144s  & 38G  & \textbf{18m} & \textbf{17G} & \textbf{63.7} \\
{EMLoC*}   & 27.6\%        & 85s   & \textbf{24G} & 19m  & \textbf{17G} & 60.9 \\
\bottomrule
\end{tabular}
\end{table}

In Section~\ref{sec:exp_ablation} and Table~\ref{tab:tab4_pruning_strategy} of the original paper, we compared our adaptive EMLoC with two static KV-cache algorithms. Table~\ref{tab:more_prune_strategy} extends this comparison (Table~\ref{tab:tab4_pruning_strategy}) by including PyramidInfer~\cite{yang2024pyramidinfer} and FastGen~\cite{ge2023model}. Most KV-cache methods focus on uni-modal text compression, but fail to maintain original performance with a high compression ratio. EMLoC retains only 22.4\% of tokens while achieving 63.7\% accuracy, outperforming FastGen (49.3\% accuracy with 36\% tokens) and PyramidInfer (55.6\% accuracy with 24.6\% tokens). Unlike existing KV-cache methods, EMLoC effectively maintains the full-context performance while significantly reducing the context length, thus improving efficiency.

To optimize the trade-off between adaptation cost and inference performance, we explore \emph{increasing the chunk number from 10 to 20 and a group-wise strategy} (every two layers share the same retention ratio). This variant, \emph{EMLoC*}, reduces adaptation time from 144s to 85s and memory from 38G to 24G, at the cost of a slight accuracy degradation(63.7→60.9) and a higher retention ratio (22.4\% → 27.6). This allows for a flexible implementation in computation constrained scenarios. The adaptation cost is significantly smaller compared with its gains in inference efficiency.

\begin{table}[t]
\small
\centering
\caption{Comparison with LongVA and MLoC on VideoMME w/o subtitles with 384 frames.}
\begin{tabular}{cccccc}
\hline
\makecell{Method} & \makecell{Context\\Length} & \makecell{LLM\\FLOPs} & \makecell{LLM\\Time} & \makecell{Peak\\Memory} & \makecell{Overall\\ACC} \\
\hline
LongVA & 55.5k & 1715.5T & 22h & 41G & 51.8 \\
MLoC   & 27.9k & 554.8T  & 7h  & 38G & \textbf{60.3} \\
EMLoC  & \textbf{2.3k} & \textbf{272.0T} & \textbf{5h} & \textbf{24G} & 60.1 \\
\hline
\end{tabular}
\end{table}
\subsection{Experiment on Long-Video Benchmark.}
In LongVA, each frame consists of 144 tokens, whereas in Qwen2-VL, 144 tokens represent every two frames through temporal pooling. Compared to our baseline MLoC, \emph{EMLoC significantly reduces computational overhead while maintaining nearly the same accuracy.} Specifically, EMLoC reduces the average context length from 27.9k to just 2.3k tokens, LLM FLOPs from 554.8T to 272.0T, inference time from 7 hours to 5 hours, and peak GPU memory from 38G to 24G, while preserving a consistent accuracy (60.1 vs. 60.3).

To achieve this efficiency, we set $\delta = 0.04$ and configure the retention ratio to: [0.02,\ 0.1,\ 0.5,\ 1.0]. Instead of optimizing the retention ratio for each layer individually (layer-wise), we adopt a \emph{group-wise strategy}, where every 14 layers are treated as a single group and share the same retention ratio. This allows for a more stable and efficient selection process during online inference. Under an identical setup (384 frames at the same resolution), both MLoC and EMLoC outperform LongVA while requiring significantly fewer computations. EMLoC also enables real-time long-video understanding on consumer-grade GPUs such as the NVIDIA 3090, making it a more practical solution for real-world applications.

\subsection{Robustness of Hyper-parameters}
Those parameters have clear meanings and are easy to adjust. For a high compression ratio, we can set a smaller retention ratio and a higher JS threshold $\delta$, and the optimal pruning strategy will be identified heuristically. Our method avoids manually adjusting numerous parameters like FastGen~\cite{ge2023model} or PyramidInfer~\cite{yang2024pyramidinfer}. Our experiments also show that the default hyperparameters are stable across different tasks, as depicted in Table~\ref{tab:sup_ablation_delta} and Table~\ref{tab:sup_ablation_retention_ratio}.

\begin{table}[h]
\small
\centering
\caption{Robustness of JS Threshold $\delta$. We show the performance under different $\delta$ values across tasks}
\label{tab:sup_ablation_delta}
\begin{tabular}{c|c|c|c}
\toprule
$\delta$ & ImageNet100 & MME-RW & OK-VQA \\
\midrule
0.002     & 63.7        & 42.2   & 58.6   \\
\textbf{0.005} & \textbf{63.7} & \textbf{42.2} & \textbf{58.7} \\
0.02      & 57.7        & 41.0   & 57.0   \\
\bottomrule
\end{tabular}
\end{table}

\begin{table}[h]
\small
\centering
\caption{Robustness of retention ratios across different tasks. We provide the performance under different retention ratios across tasks to demonstrate the robustness. }
\label{tab:sup_ablation_retention_ratio}
\begin{tabular}{c|c|c|c}
\toprule
Retention Ratios & ImageNet100 & MME-RW & OK-VQA \\
\midrule
{[0.05, 0.1, 0.2, 0.5, 1]} & 58.6 & 41.6 & 58.3 \\
\textbf{[0.1, 0.2, 0.5, 1]} & \textbf{63.7} & \textbf{42.2} & \textbf{58.7} \\
{[0.2, 0.5, 1]} & 61.6 & 41.7 & 58.6 \\
\bottomrule
\end{tabular}
\end{table}

\subsection{Impact of task disparity on EMLoC}

The impact measurement follows the method in~\cite{han2024facing}, with some modifications. Given that Qwen2VL(MLLM) is pretrained on millions of image-text pairs covering lots of vision-language tasks, we use zero-shot accuracy as an indicator of the task disparity between downstream datasets and the pretraining datasets.

MLLMs perform well on OK-VQA (52.1\%), suggesting that the data and task of OK-VQA are highly similar to those seen during pretraining. Meanwhile, ImageNet100 achieves 28\% accuracy, indicating moderate similarity. In contrast, MedXpertQA only reaches 21.5\% accuracy—near random chance in a five-choice QA—indicating significant dissimilarity.

Based on \cref{tab:sup_task_define} and \cref{tab:sup_task_disparity_performance}, the impact of task disparity can be summarized as follows:
\begin{itemize}
    \item Low task disparity (OK‑VQA, ImageNet100): When task disparity is low(highly or moderately similar to pretrained data), our EMLoC adapts well to the downstream tasks. It outperforms full fine-tuning on OK-VQA when data is limited, and achieves a average accuracy 48.2\%, which is comparable to LoRA's 47.8\%.

    \item High task disparity \& scarce data (MedXpertQA): All methods struggle. Adapting to truly novel tasks typically demands extensive continued pretraining or finetuning~\cite{lu2025fine}.

    \item Larger downstream datasets(ImageNet100 with 200 examples): Full fine‑tuning slightly outperforms both LoRA and EMLoC, echoing ~\cite{han2024facing}’s finding that “full fine‑tuning gradually closes the performance gap as dataset size grows.”

    \item Other tasks with low disparity(see \cref{tab:tab1_main}): EMLoC also facilitates on tasks with small task disparity, such as MME‑RW (OCR, remote sensing, driving), IllusionVQA (optical illusions), and YouCook2 (video captioning/activity recognition).
\end{itemize}

\begin{table}[h]
\small
\centering
\caption{Task disparity between downstream datasets and pretrained datasets.}
\label{tab:sup_task_define}
\begin{tabular}{c|c|c|c|c}
\toprule
{Dataset} & \makecell{Zero-Shot\\Accuracy} & {Task} & \makecell{Similarity to\\Pretrained Dataset} & \makecell{Task\\Disparity} \\
\hline
OK-VQA      & 52.1 & common-sense QA & Highly similar       & Low \\
ImageNet100 & 28.0 & image classification & Moderately similar  & Low \\
MedXpertQA  & 21.5 (near random) & medical QA (medical image) & Dissimilar & High \\
\bottomrule
\end{tabular}
\end{table}

\begin{table}[h]
\small
\centering
\caption{Impact of task disparity on different adaptation methods.}
\label{tab:sup_task_disparity_performance}
\begin{tabular}{c|c|c|c|c}
\toprule
{Method} & {OK-VQA} & {ImageNet100} & {MedXpertQA} & {Average} \\
\midrule
Baseline (Qwen2-VL) & 52.1 & 28.0 & 21.5 & 33.9 \\
LoRA                & \textbf{60.9} & 61.1 & 21.5 & 47.8 \\
Full Fine-tuning    & 49.7 & \textbf{64.7} & 22.0 & 45.5 \\
EMLoC               & 58.7 & 63.7 & \textbf{22.2} & \textbf{48.2} \\
\bottomrule
\end{tabular}
\end{table}

\section{Analysis of $\Delta$}
\label{sup:linear_upper_bound}
\begin{figure*}[h]
    \centering
    \includegraphics[width=1\linewidth]{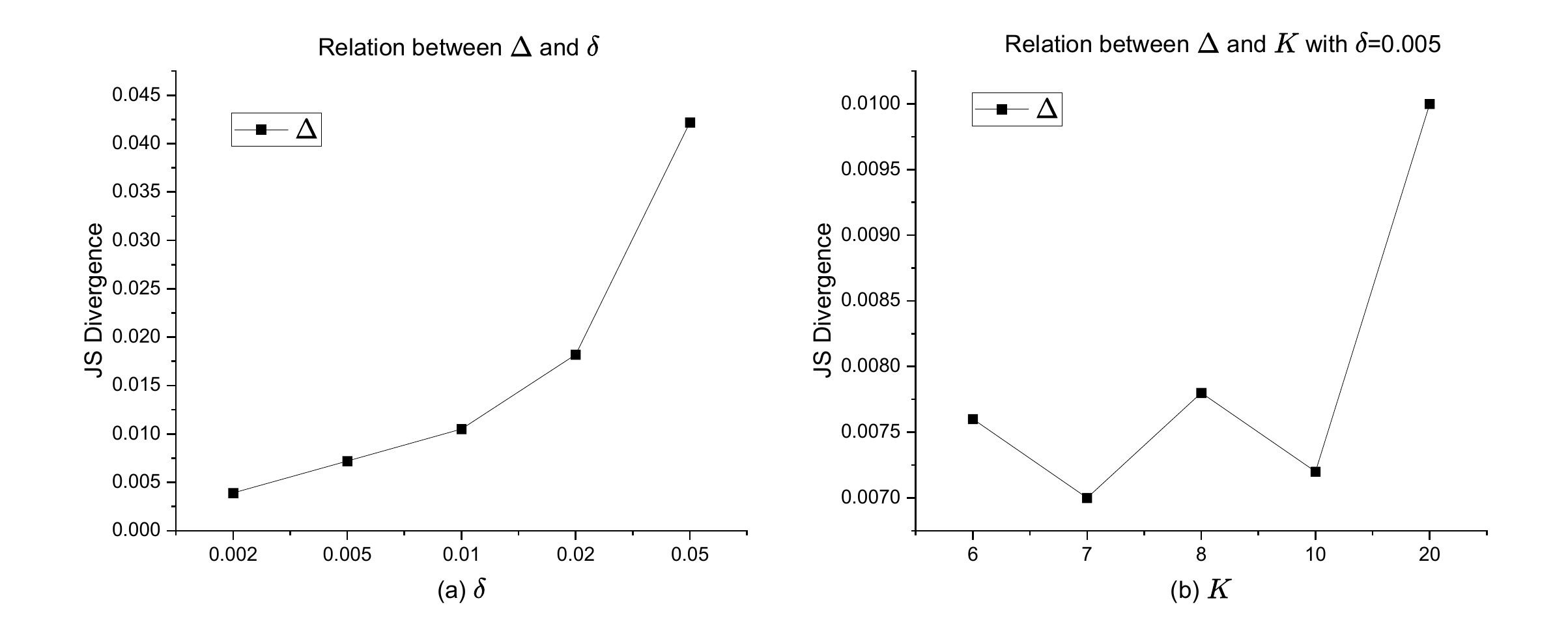}
   
    \caption{Trend of $\Delta$ on various $\delta$ and $K$.}
    \label{fig:delta_relation}
\end{figure*}

Our theoretical analysis establishes a linear relationship between the global JS distance $\sqrt{\Delta}$ and the chunk count $K$ under worst-case assumptions (\cref{eq:final_up_bound}). However, empirical observations reveal more nuanced behavior. As shown in \cref{fig:delta_relation}(a), $\Delta$ exhibits a strong positive correlation with the local constraint $\delta$, aligning with our theoretical prediction that $\sqrt{\Delta} \propto \sqrt{\delta}$. This confirms that $\delta$ effectively governs the global information loss, enabling practitioners to reliably control compression quality through this single parameter.

Notably, \cref{fig:delta_relation}(b) demonstrates that $\Delta$ remains stable within a narrow range across varying $K$ values when $\delta$ is fixed. This diverges from the theoretical upper bound's linear dependence on $K$, suggesting our worst-case analysis accommodates challenging scenarios where chunk dependencies might compound errors. In practice, however, the weak inter-chunk dependencies in real-world datasets prevent error accumulation across compression steps. Consequently, the effective upper bound simplifies to $\Delta \leq \gamma\delta$, where $\gamma$ is a small constant (typically $\gamma \leq 2$ in our experiments), rather than scaling with $K$.

This phenomenon has important practical implications: 
\begin{itemize}
    \item Hyperparameter tuning focuses primarily on $\delta$, substantially reducing configuration complexity
    \item Users can freely increase $K$ to minimize memory usage without compromising information integrity
\end{itemize}

Our method thus achieves an optimal balance between theoretical rigor and practical usability - while the theoretical bound guarantees robustness, the empirical independence between chunks enables memory-efficient compression through large $K$ values. This dual advantage makes our approach particularly suitable for long-context applications where GPU memory constraints are critical.

% to be written

\section{More Implementation Details}
\subsection{Training Details}
\label{sup:imple_details}

In \cref{tab:tab_peft_lora_ff}, we compare EMLoC with some fine-tuning methods, such as LoRA~\cite{hu2022lora}, fine-tuning, M$^2$PT~\cite{wang2024m2pt}, VPT~\cite{jia2022visual}, and E$^2$PT~\cite{han2023e2vpt}. 

In LoRA adaptation, we apply LoRA adapters to all linear modules of the LLM, including qkv\_proj, out\_proj, up\_proj, and down\_proj, while keeping the vision encoder and multi-modal projector frozen.  The rank and alpha are set to 16 and 32, respectively. In full fine-tuning, only the LLM is fine-tuned with  DeepSpeed ZeRO-3, leaving other parameters frozen. Other unspecified settings follow the default configurations in LLaMAFactory. The detailed hyperparameters are reported in \cref{tab:lora_hyper} and \cref{tab:sft_hyper}.
\begin{table}[h]
    \centering
    \caption{Hyperparameters for LoRA training} \label{tab:lora_hyper}
    \begin{tabular}{c|c}
    \toprule
    \bf  Hyperparameter    &  \bf Value \\ \midrule
     Optimizer    &  AdamW \\
     learning rate &  3e-5 \\
     batch size & 8 \\
     warmup ratio & 0.1 \\ 
     epochs & 5 \\
     clip norm & 1.0 \\ \bottomrule
    \end{tabular}
\end{table}

\begin{table}[h]
    \centering
    \caption{Hyperparameters for full fine-tuning} \label{tab:sft_hyper}
    \begin{tabular}{c|c}
    \toprule
    \bf  Hyperparameter    &  \bf Value \\ \midrule
     Optimizer    &  AdamW \\
     learning rate &  1e-5 \\
     batch size & 8 \\
     warmup ratio & 0.1 \\ 
     epochs & 1 \\
     clip norm & 1.0 \\ \bottomrule
    \end{tabular}
\end{table}

For other PEFT methods, following the default configuration in M²PT~\cite{wang2024m2pt}, the number of visual prompts and textual prompts is 20 and 10, respectively. The learning rate is set to 7e-4. The number of KV prompt token is 5. For Imagenet100, we optimize 5 epochs with 125 steps. For MME-RW and OK-VQA, we just fine-tune 25 steps.

\subsection{Dataset Details}
We evaluate our method on \textit{ImageNet100}, a subset of ImageNet-1k~\cite{deng2009imagenet} with the first 100 classes for inference efficiency. Demonstration examples are uniformly sampled from the training set, ensuring even distribution per class. For instance, in the 200-example setting, each class contributes two examples. Evaluation is conducted on the full validation set with 5000 images.
Additionally, we evaluate on several other benchmarks: \textit{ScreenSpot}\cite{cheng2024seeclick} for GUI grounding across diverse platforms,
\textit{MME-RW}\cite{zhang2024mme} for real-world tasks such as OCR, remote sensing, and autonomous driving,
\textit{IllusionvQA}\cite{shahgir2024illusionvqa} for evaluating optical-illusion understanding,
\textit{OK-VQA}\cite{marino2019ok} for open-ended question answering using external knowledge,
and \textit{YouCook2}~\cite{zhou2018towards} for cooking-video-related tasks, e.g., captioning or activity recognition.

\end{document}